\documentclass{article}
\usepackage{ijcai22}

\usepackage{times}
\usepackage{soul}
\usepackage{url}
\usepackage[hidelinks]{hyperref}
\usepackage[utf8]{inputenc}
\usepackage[small]{caption}
\usepackage{subcaption}
\usepackage{graphicx}
\usepackage{amsmath}
\usepackage{amsthm}
\usepackage{bbm}
\usepackage{bm}
\usepackage{booktabs}
\usepackage{algorithm}
\usepackage{algorithmic}
\usepackage{csquotes}
\newcommand{\norm}[1]{\left\lVert#1\right\rVert}

\title{Dynamic Sparse Training for Deep Reinforcement Learning}

\author{
Ghada Sokar$^1$
\and
Elena Mocanu$^2$\and
Decebal Constantin Mocanu$^{1,2}$\and
Mykola Pechenizkiy$^1$\And\\
Peter Stone$^3$
\affiliations
$^1$Eindhoven University of Technology, The Netherlands\\
$^2$University of Twente, The Netherlands\\
$^3$The University of Texas at Austin, Sony AI, United States
\emails
\{g.a.z.n.sokar, m.pechenizkiy\}@tue.nl,
\{e.mocanu, d.c.mocanu\}@utwente.nl,
pstone@cs.utexas.edu
}

\begin{document}

\maketitle

\begin{abstract}
Deep reinforcement learning (DRL) agents are trained through trial-and-error interactions with the environment. This leads to a long training time for dense neural networks to achieve good performance. Hence, prohibitive computation and memory resources are consumed. Recently, learning efficient DRL agents has received increasing attention. Yet, current methods focus on accelerating \textit{inference} time. In this paper, we introduce for the \textit{first time} a dynamic sparse training approach for deep reinforcement learning to accelerate the \textit{training} process. The proposed approach trains a \textit{sparse} neural network from scratch and dynamically adapts its topology to the changing data distribution during training. Experiments on continuous control tasks show that our \textit{dynamic sparse} agents achieve higher performance than the equivalent dense methods, reduce the parameter count and floating-point operations (FLOPs) by 50\%, and have a faster learning speed that enables reaching the performance of dense agents with $40-50$\% reduction in the training steps\footnote{Code is available at: https://github.com/GhadaSokar/Dynamic-Sparse-Training-for-Deep-Reinforcement-Learning.\\Proceedings of the 31st International Joint Conference on Artificial Intelligence (IJCAI-22).}.
\end{abstract}

\section{Introduction}
Deep reinforcement learning (DRL) has achieved remarkable success in many applications. The power of deep neural networks as function approximators allows RL agents to scale to environments with high-dimensional state and action spaces. This enables high-speed growth in the field and the rise of many methods that improve the performance and stability of DRL agents \cite{wang2020deep}. While the achieved performance is impressive, a long training time is required to obtain this performance. For instance, it took more than 44 days to train a Starcraft II agent using 32 third-generation tensor processing units (TPUs) \cite{vinyals2019grandmaster}. The very long training time leads to high energy consumption and prohibitive memory and computation costs. In this paper, we ask the following question: \textit{Can we provide efficient DRL agents with less computation cost and energy consumption while maintaining superior performance?}

Few recent works attempt to accelerate the \textit{inference} time of DRL agents via pruning \cite{livne2020pops} or training a compact network under the guidance of a larger network (knowledge distillation) \cite{zhang2019accelerating}. Despite the computational improvement achieved at inference, extensive computations throughout the training of \textit{dense} networks are still consumed. Our goal is to accelerate the training process as well as the inference time of DRL agents. 

The long training time of a DRL agent is due to two main factors: \textbf{(1)} the extensive computational cost of training deep neural networks caused by the very high number of network parameters \cite{jouppi2017datacenter} and \textbf{(2)} the learning nature of a DRL agent in which its policy is improving through many trial-and-error cycles while interacting with the environment and collecting a large amount of data. In this paper, we introduce dynamic sparse training (DST) ~\cite{mocanu2021sparse,hoefler2021sparsity} in the DRL paradigm for the first time to address these two factors. Namely, we propose an efficient training approach that can be integrated with existing DRL methods. Our approach is based on training \textit{sparse} neural networks from scratch with a fixed parameter count throughout training \textbf{(1)}. During training, the sparse topology is optimized via adaptation cycles to \textit{quickly adapt} to the online changing distribution of the data \textbf{(2)}. Our training approach enables reducing memory and computation costs substantially. Moreover, the quick adaptation to the new samples from the improving policy during training leads to a faster learning speed.

In fact, the need for neural networks that can adapt, e.g., change their control policy dynamically as environmental conditions change, was broadly acknowledged by the RL community~\cite{Stanley03evolvingadaptive}. Although prior works related to the automatic selection of function approximation based on neuroevolution exist~\cite{HEIDRICHMEISNER2009152}, perhaps the most connected in the spirit to our proposed method is a combination of NeuroEvolution of Augmenting Topologies (NEAT) \cite{stanley2002evolving} and temporal difference (TD) learning (i.e., NEAT+Q \cite{whiteson2006evolutionary}). Still, the challenge remains, and cutting-edge DRL algorithms do not account for the benefits of adaptive neural networks training yet.

Our contributions in this paper are as follows:
\begin{itemize}
    \item The principles of dynamic sparse training are introduced in the deep reinforcement learning field for the first time.
    \item Efficient improved versions of two state-of-the-art algorithms, TD3 \cite{fujimoto2018addressing} and SAC \cite{haarnoja2018soft}, are obtained by integrating our proposed DST approach with the original algorithms.
    \item Experimental results show that our training approach reduces the memory and computation costs of training DRL agents by 50\% while achieving superior performance. Moreover, it achieves a faster learning speed, reducing the required training steps.
    \item Analysis insights demonstrate the promise of dynamic sparse training in advancing the field and allowing for DRL agents to be trained and deployed on low-resource devices (e.g., mobile phones, tablets, and wireless sensor nodes) where the memory and computation power are strictly constrained. See Appendix \ref{appendix:HW_SW} for discussion. 
\end{itemize}

\section{Related Work}
\textbf{Sparsity in DRL.} To the best of our knowledge, the current advance in deep reinforcement learning is achieved using \textit{dense} neural networks. 
Few recent studies have introduced sparsity in DRL via pruning. PoPS \cite{livne2020pops} first trains a dense teacher neural network to learn the policy. This dense teacher policy network guides the iterative pruning and retraining of a student policy network via knowledge distillation. In \cite{zhang2019accelerating}, the authors aim to accelerate the behavior policy network and reduce the time for sampling. They use a smaller network for the behavior policy and learn it simultaneously with a large dense target network via knowledge distillation. GST \cite{lee2021gst} was proposed as an algorithm for weight compression in DRL training by simultaneously utilizing weight grouping and pruning. Some other works \cite{yu2019playing,vischer2021lottery} studied the existence of the lottery ticket hypothesis \cite{frankle2018lottery} in RL, which shows the presence of sparse subnetworks that can outperform dense networks when they are trained from scratch. Pruning dense networks increases the computational cost of the training process as it requires iterative cycles of pruning and retraining \cite{molchanov2019pruning,renda2020comparing,molchanov2019importance,chen2021lottery}. This work introduces the first efficient training algorithm for DRL agents that trains sparse neural networks directly from scratch and adapts to the changing distribution. 

\textbf{Dynamic Sparse Training (DST).}
DST is the class of algorithms that train sparse neural networks \textit{from scratch} and jointly optimize the weights and the sparse topology during training. This direction aims to reduce the computation and memory overhead of training dense neural networks by leveraging the redundancy in the parameters (i.e., being over-parametrized) ~\cite{denil2013predicting}. Efforts in this line of research are devoted to supervised and unsupervised learning. The first work in this direction was proposed by \cite{mocanu2018scalable}. They proposed a Sparse Evolutionary Training algorithm (SET) that dynamically changes the sparse connectivity during training based on the values of the connections. The method achieves higher performance than dense models and static sparse neural networks trained from scratch. The success of the SET algorithm opens the path to many interesting DST methods that bring higher performance gain. These algorithms differ from each other in the way the sparse topology is adapted during training \cite{mostafa2019parameter,evci2020rigging,dettmers2019sparse,jayakumar2020top,bellec2018deep,pmlr-v139-liu21p,NEURIPS2020_b4418237,yuan2021mest}. DST demonstrated its success in other fields as well, such as feature selection \cite{atashgahi2020quick}, continual learning \cite{SOKAR20211}, ensembling \cite{liu2021deep}, federated learning \cite{zhu2019multi}, text classification and language modeling tasks \cite{liu2021efficient}, and adversarial training \cite{ozdenizci2021training}. 

In this work, we adopt the topological adaptation from the SET method in our proposed approach. The motivation is multifold. First, SET is simple yet effective; it achieves the same or even higher accuracy than dense models with high sparsity levels across different architectures (e.g., multi-layer perceptrons, convolutional neural networks, restricted Boltzmann machines). Second, unlike other DST methods that use the values of non-existing (masked) weights in the adaptation process, SET uses only the values of existing sparse connections. This makes SET truly sparse and memory-efficient \cite{onemillionneurons}. Finally, it does not need high computational resources for the adaptation process. It uses readily available information during the standard training. These factors are favorable for our goal to train efficient DRL agents suitable for real-world applications. We leave evaluating other topological adaptation strategies for future work.  

\section{Proposed Method}
\label{method}
In this section, we describe our proposed method, which introduces dynamic sparse training for the DRL paradigm. Here, we focus on integrating our training approach with one of the state-of-the-art DRL methods; Twin Delayed Deep Deterministic policy gradient (TD3) \cite{fujimoto2018addressing}. We named our new approach Dynamic Sparse TD3 or \enquote{DS-TD3} for short. TD3 is a popular and efficient DRL method that offers good performance in many tasks \cite{joshi2021application,shi2020adaptive,ye2021real,woo2020real,hou2021control}. Yet, our approach can be merged into other DRL algorithms as well. Appendix \ref{appendix:DS-SAC} shows the integration with soft actor-critic (SAC) \cite{haarnoja2018soft}.

TD3 is an actor-critic method that addresses the overestimation bias in previous actor-critic approaches. In actor-critic methods, a policy $\pi$ is known as the \textit{actor}, and a state-value function $Q$ is known as the \textit{critic}. Target networks are used to maintain fixed objectives for the actor and critic networks over multiple updates. In short, TD3 limits the overestimation bias using a pair of critics. It takes the smallest value of the two critic networks to estimate the $Q$ value to provide a more stable approximation. To increase the stability, TD3 proposed a delayed update of the actor and target networks. In addition, the weights of the target networks are slowly updated by the current networks by some proportion $\tau$. \textit{In this work}, we aim to dynamically train the critics and actor networks along with their corresponding target networks from scratch with sparse neural networks to provide efficient DRL agents. In the rest of this section, we will explain our proposed DST approach for TD3. The full details are also provided in Algorithm \ref{alg:DS_TD3}.

Our proposed DS-TD3 consists of four main phases: sparse topology initialization, adaptation schedule, topological adaptation, and maintaining sparsity levels. 

\textbf{Sparse Topology Initialization (Algorithm \ref{alg:DS_TD3} \textbf {L1-L4}).} TD3 uses two critic networks ($Q_{\bm{\theta}_1}, Q_{\bm{\theta}_2}$) and one actor network ($\pi_{\bm{\phi}}$) parameterized by $\bm{\theta_{1}}=\{\bm{\theta_{1}}^l\}|_{l=1}^L$, $\bm{\theta_{2}}=\{\bm{\theta_{2}}^l\}|_{l=1}^L$, and $\bm{\phi}=\{\bm{\phi}^l\}|_{l=1}^L$ respectively; where $L$ is the number of layers in a network. We initialize each of the actor and critic networks with a sparse topology. Sparse connections are allocated in each layer between the hidden neurons at layer $l-1$ and layer $l$. We represent the locations of the sparse connections by a binary mask $\bm{M}=\{\bm{M}^l\}|_{l=1}^L$. Following \cite{mocanu2018scalable}, we use Erdős–Rényi random graph \cite{erdos1960evolution} to initialize a sparse topology in each layer $l$. Namely, the probability of a connection $j$ in layer $l$ is given by:
\begin{equation}
    p(\bm{M}^{j}) = \lambda^{l} \frac{n^{l}+n^{l-1}}{n^{l} \times n^{l-1}},
\end{equation}
where $\lambda^{l}$ is a hyperparameter to control the sparsity level in layer $l$, and $n^{l-1}$ and $n^l$ are the neurons count in layers $l-1$ and $l$, respectively. $\bm{M}^{j} \in \{0,1\}$; a value of 1 means the existence of a weight in location $j$. We omit the index $l$ from the mask and weight matrices for readability. A sparse topology is created in each layer for the actor and critic networks:
\begin{equation}
 \begin{gathered}
 \begin{aligned}
   \bm{\phi} &= \bm{\phi} \odot \bm{M_{\phi}},\\
   \bm{\theta_{i}} &= \bm{\theta_{i}} \odot \bm{M_{\theta_i}}, \qquad \forall i \in\{1,2\},
   \end{aligned}
  \end{gathered}
\end{equation}
where $\odot$ is an element-wise multiplication operator and $ \bm{M_{\phi}}$, $\bm{M_{\theta_1}}$, and $\bm{M_{\theta_2}}$ are binary masks to represent the sparse weights in the actor and two critic networks, respectively. The initial sparsity level is kept fixed during the training. 

The target policy and target critic networks are parameterized by $\bm{\phi'}$, $\bm{\theta'}_{1}$, and $\bm{\theta'}_{2}$, respectively. Initially, the target networks have the same sparse topology and the same weights as the current networks: $\bm{\phi'} \leftarrow \bm{\phi}$, $\bm{\theta'}_1 \leftarrow \bm{\theta_1}$, $\bm{\theta'}_2 \leftarrow \bm{\theta_2}$. 

After the topological and weight initialization, the agent collects enough data before training using a purely exploratory policy. During training, for each time step, TD3 updates the pair of critics towards the minimum target value of actions selected by the target policy $\pi_{\bm{\phi'}}$:
\begin{equation}
y = r + \gamma \min_{i=1,2} Q_{\bm{\theta'}_i}(s', \pi_{\bm{\phi'}}(s') + \epsilon), \\
\end{equation}
where $\gamma$ is the discounting factor, $r$ is the current reward, $s'$ is the next state, and $\epsilon \sim clip(\mathcal{N}(0, \tilde \sigma), -c, c) $ is the proposed clipped noise by TD3, defined by $\tilde \sigma$, to increase the stability; where $c$ is the clipped value. As discussed, TD3 proposed to delay the update of the policy network to first minimize the error in the value network before introducing a policy update. Therefore, the actor network is updated every $d$ steps with respect to $Q_{\theta1}$ as shown in Algorithm \ref{alg:DS_TD3} \textbf{L17-L19}.

During the weight optimization of the actor and critic networks, the values of the existing sparse connections are only updated (i.e., the sparsity level is kept fixed). The sparse topologies of the networks are also optimized during training according to our proposed adaptation schedule.
\begin{algorithm}[tb]
\caption{DS-TD3 ($\lambda^{l}$, $\eta$, $e$, $N$, $\tau$, $d$)}
\label{alg:DS_TD3}
 \begin{algorithmic}[1]
%    \INPUT Horizon $T$, batch size $N$, exploration noise $\s_\textnormal{exploration}$, target network update rate $\tau$, target policy noise $\s_\textnormal{target}$, policy delay $d$. 
   %\STATE \textbf{Require:} $\lambda^{l}$, $\eta$, $e$, batch size $N$
   \STATE Initialize critic networks $Q_{\bm{\theta_1}}$, $Q_{\bm{\theta_2}}$ and actor network $\pi_{\bm{\phi}}$ with sparse parameters $\bm{\theta_1}$, $\bm{\theta_2}$, $\bm{\phi}$ with a sparsity level defined by $\lambda^{l}$:
   \STATE Create $\bm{M_{\phi}}$, $\bm{M_{\theta_1}}$, and $\bm{M_{\theta_2}}$ with Erdős–Rényi graph
   \STATE $\bm{\theta_{1}} \leftarrow$ $\bm{\theta_{1}} \odot$ $\bm{M_{\theta_1}}$, $\bm{\theta_{2}} \leftarrow$ $\bm{\theta_{2}} \odot$ $\bm{M_{\theta_2}}$, $\bm{\phi} \leftarrow \bm{\phi} \odot \bm{M_{\phi}}$
   \STATE Initialize target networks $\bm{\theta'}_1 \leftarrow \bm{\theta_1}$, $\bm{\theta'}_2 \leftarrow \bm{\theta_2}$, $\bm{\phi'} \leftarrow \bm{\phi}$
   \STATE Initialize replay buffer $\mathcal{B}$
   \FOR{$t=1$ {\bfseries to} $T$}
   \STATE Select action with exploration noise $a \sim \pi_{\bm{\phi}}(s) + \epsilon$, 
   \STATE $\epsilon \sim \mathcal{N}(0, \sigma)$ and observe reward $r$ and new state $s'$
   \STATE Store transition tuple $(s, a, r, s')$ in $\mathcal{B}$ 
   \STATE Sample mini-batch of $N$ transitions from $\mathcal{B}$
   \STATE $\tilde a \leftarrow \pi_{\bm{\phi'}}(s') + \epsilon, \quad \epsilon \sim clip(\mathcal{N}(0, \tilde \sigma), -c, c)$
   %\STATE Use Clipped Double Q-learning target:
   \STATE $y \leftarrow r + \gamma \min_{i=1,2} Q_{\bm{\theta'}_i}(s', \tilde a)$
   %\STATE Update $\theta_i$ to minimize $N^{-1} \sum (y - Q_{\theta_i}(s,a))^2$
   %$\{Q_{\theta_i}\}_{i=1}^2$
   \STATE $\bm{\theta_i} \leftarrow argmin_{\bm{\theta_i}} \frac{1}{N} \sum (y - Q_{\bm{\theta_i}}(s,a))^2$
    \IF{$t$ mod $e$} 
      \STATE $\bm{\theta_i}\leftarrow$ TopologicalAdaptation$(\bm{\theta_i}, \bm{M_{\theta_{i}}},\eta)$ (Algo. \ref{alg:topo_evolution})
   \ENDIF
   \IF{$t$ mod $d$}
   \STATE Update $\bm{\phi}$ by the deterministic policy gradient:
   \STATE $\nabla_{\bm{\phi}} J(\bm{\phi}) \leftarrow \frac{1}{N} \sum \nabla_{a} Q_{\bm{\theta_1}}(s, a)|_{a=\pi_{\bm{\phi}}(s)} \nabla_{\bm{\phi}} \pi_{\bm{\phi}}(s)$
   \IF{$t$ mod $e$}
    \STATE $\bm{\phi} \leftarrow$ TopologicalAdaptation$(\bm{\phi}, \bm{M_{\phi}},\eta)$ (Algo. \ref{alg:topo_evolution})
   \ENDIF
   \STATE Update target networks:
   \STATE $\bm{\theta'}_i \leftarrow \tau \bm{\theta_i} + (1 - \tau) \bm{\theta'}_i$
   \STATE $\bm{\phi'} \leftarrow \tau \bm{\phi} + (1 - \tau) \bm{\phi'}$
      \STATE $\bm{\theta'_i} \leftarrow$ MaintainSparsity$(\bm{\theta'}_i,\norm{\bm{\theta_i}}_0)$ (Algo. \ref{alg:maint_sparsity})
    \STATE $\bm{\phi'} \leftarrow$ MaintainSparsity$(\bm{\phi'}, \norm{\bm{\phi}}_0)$  (Algo. \ref{alg:maint_sparsity})
   \ENDIF
   \ENDFOR
\end{algorithmic}
\end{algorithm}
\begin{algorithm}[tb]
\caption{Topological Adaptation ( $\bm{X}$, $\bm{M}$, $\eta$)}
\label{alg:topo_evolution}
 \begin{algorithmic}[1]
%\STATE \textbf{Require:} $\bm{X}$, $\bm{M}$, $\eta$
    \STATE $c \leftarrow \eta \norm{\bm{X}}_{0}$
    \STATE $c_p \leftarrow c/2$ ; \quad $c_n \leftarrow c/2$
    \STATE $\tilde{\bm{X}}^p \leftarrow$ get\_\textit{$c_{p}$-{th}}\_smallest\_positive$(\bm{X})$
    \STATE $\tilde{\bm{X}}^n \leftarrow$ get\_\textit{$c_{n}$-{th}}\_largest\_negative$(\bm{X})$
    %\STATE Drop least important connections:
    \STATE $\bm{M}^j \leftarrow \bm{M}^j -  \mathbbm{1} [(0<\bm{X}^j< \tilde{\bm{X}}^p) \lor (0>\bm{X}^j>\tilde{\bm{X}}^n)]$
    %\STATE Grow new connections:
    \STATE Generate $c$ random integers $\{x\}|_1^c$ 
    \STATE $\bm{M}^j \leftarrow \bm{M}^j +  \mathbbm{1} [(j==x) \land (\bm{X}^j==0)]$
    \STATE $\bm{X} \leftarrow \bm{X} \odot \bm{M}$
\end{algorithmic}
\end{algorithm}
\begin{algorithm}[tb]
\caption{Maintain Sparsity ($\bm{X}$, $k$)}
\label{alg:maint_sparsity}
 \begin{algorithmic}[1]
 %\STATE \textbf{Require:} $\bm{X}$, $k$
   \STATE $\tilde{\bm{X}}$ $\leftarrow$ Sort\_Descending$(|X|)$ 
   \STATE  $\bm{M}^j = \mathbbm{1} [|\bm{X}^j| - \tilde{\bm{X}}^k\geq0], \forall j \in\{1,... \norm{\bm{X}}_0\}$
    \STATE $\bm{X} = \bm{X} \odot \bm{M}$
\end{algorithmic}
\end{algorithm}

\textbf{Adaptation Schedule.} The typical practice in DST methods applied in the supervised setting is to perform the dynamic adaptation of the sparse topology after each training epoch. However, this would not fit the RL setting directly due to its dynamic learning nature. In particular, an RL agent faces instability during training due to the lack of a true target objective. The agent learns through trial and error cycles, collecting the data online while interacting with the environment. Adapting the topology very frequently in this learning paradigm would limit the exploration of effective topologies for the data distribution and give a biased estimate of the current one.  To address this point, we propose to delay the adaptation process and perform it every $e$ time steps, where $e$ is a hyperparameter. This would allow the newly added connections from the previous adaptation process to grow. Hence, it would also give better estimates of the connections with the least influence in the performance and an opportunity to explore other effective ones. Analysis of the effect of the adaptation schedule in the success of applying dynamic sparse training in the RL setting is provided in Section \ref{sec:analysis}.

\textbf{Topological Adaptation (Algorithm \ref{alg:topo_evolution}).} We adopt the adaptation strategy of the SET method \cite{mocanu2018scalable} in our approach. The sparse topologies are optimized according to our adaptation schedule. Every $e$ steps, we update the sparse topology of the actor and critic networks. Here, we explain the adaptation process on the actor network as an example. The same strategy is applied for the critic networks. 

The adaptation process is performed through a \enquote{drop-and-grow} cycle which consists of two steps. \textbf{The first step} is to \textit{drop} a fraction $\eta$ of the least important connections from each layer. This fraction is a subset ($c_p$) of the smallest positive weights and a subset ($c_n$) of the largest negative weights. Thus, the removed weights are the ones closest to zero. Let $\tilde{\bm{\phi}}^p$ and $\tilde{\bm{\phi}}^n$ be the $c_p$-th smallest positive and the $c_n$-th largest negative weights, respectively. The mask $\bm{M_{\phi}}$ is updated to represent the dropped connections as follows:
\begin{equation}
\resizebox{.91\linewidth}{!}{$
\centering
 \bm{M^j_{\phi}} = \bm{M^j_{\phi}} -  \mathbbm{1} [(0<\bm{\phi}^j< \tilde{\bm{\phi}}^p) \lor (0>\bm{\phi}^j>\tilde{\bm{\phi}}^n)], \quad \forall j \in\{1,...,\norm{\bm{\phi}}_0\},
$}
\end{equation}
where $\bm{M^j_{\phi}}$ is the element $j$ in $\bm{M_{\phi}}$, $\mathbbm{1}$ is the indicator function, $\lor$ is the logical OR operator, and $\norm{.}_0$ is the standard $L_0$ norm. \textbf{The second step} is to \textit{grow} the same fraction $\eta$ of removed weights in random locations from the non-existing weights in each layer. $\bm{M_{\phi}}$ is updated as follows: 
\begin{equation}
\resizebox{.91\linewidth}{!}{$
\centering
   \bm{M^j_{\phi}} = \bm{M^j_{\phi}} +  \mathbbm{1} [(j==x) \land (\bm{\phi}^j==0)] , \quad \forall j \in\{1,..,\norm{\bm{\phi}}_0\},
$}
\end{equation}
where $x$ is a random integer generated from the discrete uniform distribution in the interval $[1, n^{(l-1)}\times(n^l)]$ and $\land$ is the logical AND operator. The weights of the newly added connections are zero-initialized ($\bm{\phi} = \bm{\phi} \odot \bm{M_{\phi}}$).

\textbf{Maintain Sparsity Level in Target Networks (Algorithm \ref{alg:maint_sparsity}).} 
TD3 delays the update of the target networks to be performed every $d$ steps. In addition, the target networks are slowly updated by some proportion $\tau$ instead of making the target networks exactly match the current ones (Algorithm \ref{alg:DS_TD3} \textbf{L23-L25}). These two points lead to a slow deviation of the sparse topologies of the target networks from current networks. Consequently, the slow update of the target networks by $\tau$ would slowly increase the number of non-zero connections in the target networks over time. \textit{To address this}, after each update of the target networks, we prune the extra connections that make the total number of connections exceed the initial defined one. We prune the extra weights based on their smallest magnitude. Assume we have to retain $k$ connections. The target masks of the actor ($\bm{M'}_{\bm{\phi'}}$) and critics ($\bm{M'}_{\bm{\theta'}_1}$, $\bm{M'}_{\bm{\theta'}_2}$) are calculated as follows: 
\begin{equation}
\resizebox{.91\linewidth}{!}{$
\begin{gathered}
    \begin{aligned}
    \bm{M'}^j_{\bm{\phi'}} &= \mathbbm{1} [|\bm{\phi'}^j| - \tilde{\bm{\phi'}}^k\geq0], \quad \forall j \in\{1,...,\norm{\bm{\phi'}}_0\},\\
    \bm{M'}^j_{\bm{\theta'}_i} &= \mathbbm{1} [|\bm{\theta'}^j_i| - \tilde{\bm{\theta'}}^k_i\geq0], \quad \forall j \in\{1,...,\norm{\bm{\theta'}_i}_0\}, \quad \forall i \in\{1,2\},  
    \end{aligned}
\end{gathered}
$}
\end{equation}
where $\tilde{\bm{\phi'}}^k$ and $\tilde{\bm{\theta'}_i}^k$ is the $k$-th largest magnitude in the actor and critics respectively, and  $|(.)^j|$ is the magnitude of element $j$ in the matrix. The target networks are updated as follows:
\begin{equation}
    \begin{gathered}
    \begin{aligned}
        \bm{\phi'} &= \bm{\phi'} \odot \bm{M'}_{\bm{\phi'}}, \\
        \bm{\theta'}_i &= \bm{\theta'}_i \odot \bm{M'}_{\bm{\theta'}_i} \quad \forall i \in\{1,2\}. 
    \end{aligned}
    \end{gathered}
\end{equation}
\begin{figure*}
 \centering
 \begin{subfigure}{0.55\columnwidth}
    \centering
    \includegraphics[width=\linewidth]{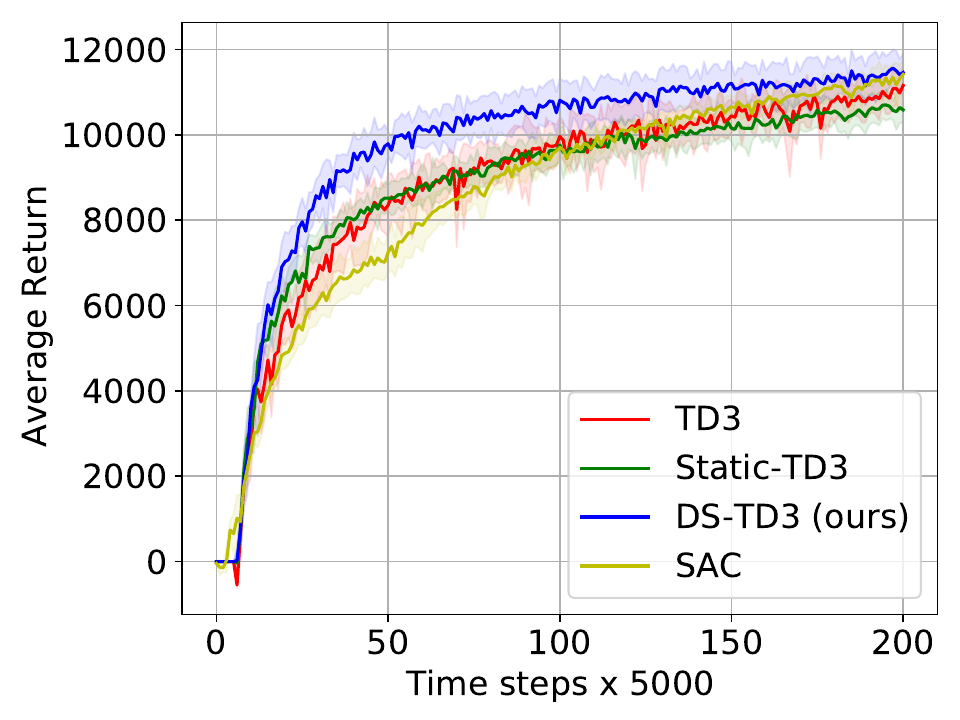}
      \caption{HalfCheetah-v3.}
 \end{subfigure}
 \begin{subfigure}{0.55\columnwidth}
    \centering
    \includegraphics[width=\linewidth]{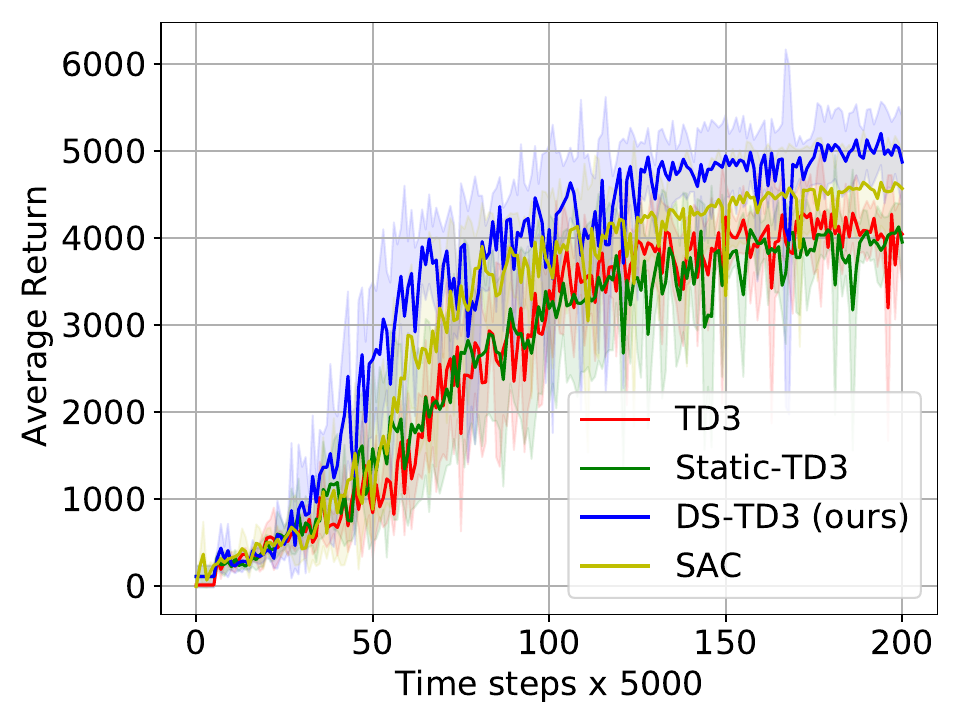}
      \caption{Walker2d-v3.}
 \end{subfigure}
\begin{subfigure}{0.55\columnwidth}
    \centering
    \includegraphics[width=\linewidth]{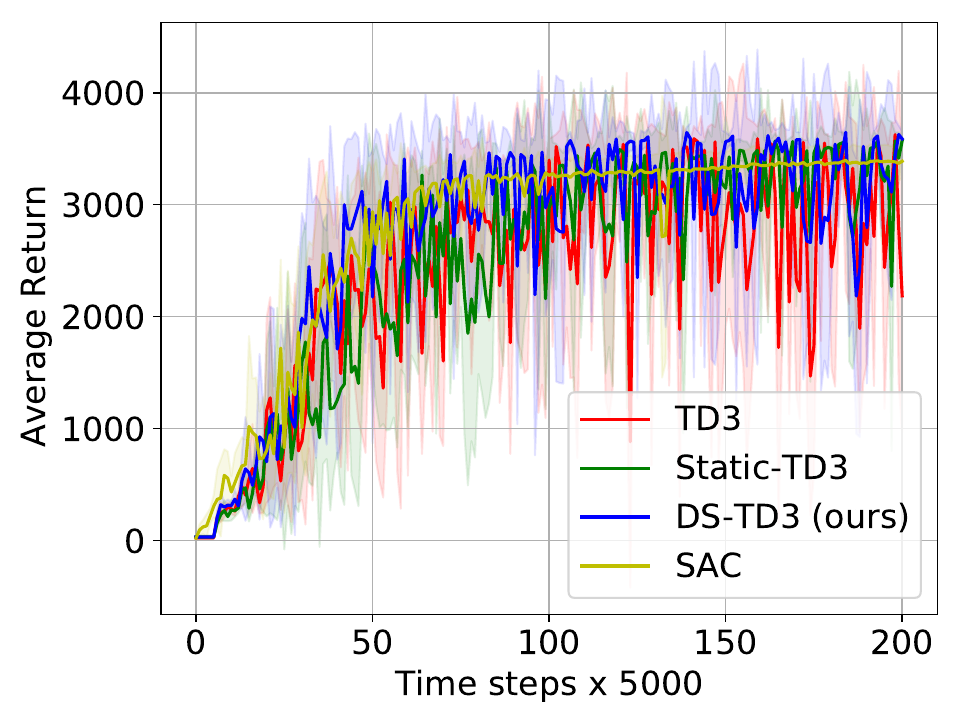}
      \caption{Hopper-v3.}
 \end{subfigure}
 \begin{subfigure}{0.55\columnwidth}
    \centering
    \includegraphics[width=\linewidth]{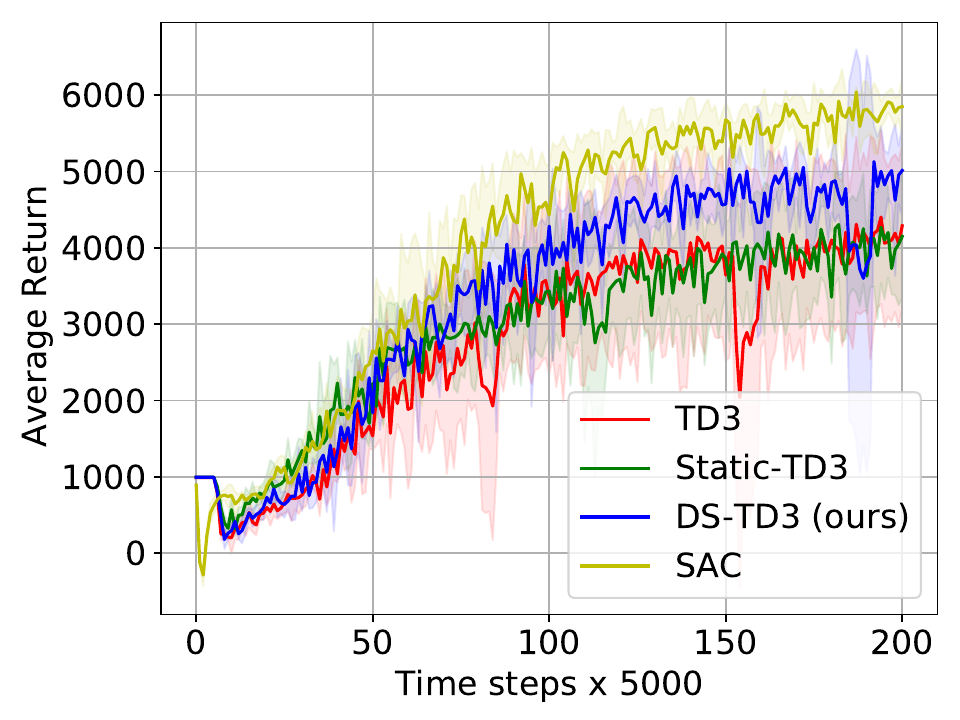}
      \caption{Ant-v3.}
 \end{subfigure}
  \begin{subfigure}{0.55\columnwidth}
    \centering
    \includegraphics[width=\linewidth]{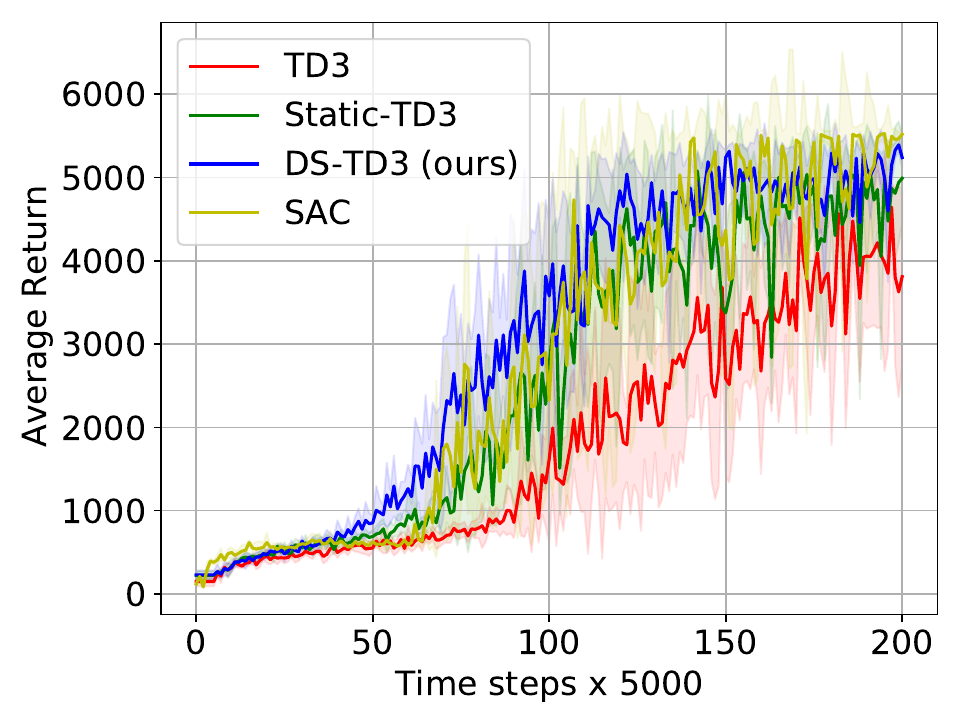}
      \caption{Humanoid-v3.}
 \end{subfigure}
\caption{Learning curves of the studied algorithms on different continuous control tasks. The shaded region represents the standard deviation of the average evaluation over 5 runs.} 
\label{learning_curves}
\end{figure*}
\section{Experiments and Results}
\label{sec:experiments}
In this section, we assess the efficiency of our proposed dynamic sparse training approach for the DRL paradigm and compare it to state-of-the-art algorithms. Experimental settings are provided in Appendix \ref{appendix:Experimental_Details}.

\textbf{Baselines.} We compare our proposed DS-TD3 against the following baselines: (1) \textit{TD3}~\cite{fujimoto2018addressing}, the original TD3 where dense networks are used for actor and critic models, (2) \textit{Static-TD3}, a variant of TD3 where the actor and critic models are initialized with sparse neural networks which are kept fixed during training (i.e., there is no topological optimization), and (3) \textit{SAC}~\cite{haarnoja2018softapp}, a popular off-policy algorithm in which the policy is trained to maximize a trade-off between expected return and entropy which results in policies that explore better. 

\textbf{Benchmarks.} We performed our experiments on MuJoCo continuous control tasks, interfaced through OpenAI Gym. We evaluate our proposed approach on five challenging environments (HalfCheetah-v3, Hopper-v3, Walker2d-v3, Ant-v3, and Humanoid-v3).

\textbf{Metrics.} We use multiple metrics to assess the efficiency of the studied DRL methods (Details are in Appendix \ref{appendix:evaluation_metrics}): \textbf{(1) Return} which is the standard metric used in DRL to measure the \textit{performance} of an agent, \textbf{(2) Learning curve area (LCA)} which estimates the \textit{learning speed} of a model (i.e., how quickly a model learns) \cite{chaudhry2018efficient} by measuring the area under the training curve of a method, \textbf{(3) Network size (\#params)} which measures the \textit{memory cost} via the number of network parameters, and \textbf{(4) Floating-point operations (FLOPs)} which estimate the \textit{computational cost} required for training. It is the typical metric in the literature to compare a DST method against its dense counterpart (see Appendix \ref{appendix:evaluation_metrics} for discussion). 

\subsection{Results}
\label{sec:results}
\textbf{Learning Behavior and Speed.} Figure \ref{learning_curves} shows the learning curve of studied methods. DS-TD3 has a much faster learning speed than the baselines, especially at the beginning of the training. After 40-50\% of the steps, DS-TD3 can achieve the final performance of TD3. Static-TD3 does not have this favorable property which reveals the importance of optimizing the sparse topology during training to adapt to the incoming data. The learning behavior of DS-TD3 is also faster than SAC in all environments except one. Table \ref{table:learning_speed} shows the learning curve area (LCA) of each method. DS-TD3 has higher LCA than TD3 and static-TD3 in all environments. It is also higher than SAC in three environments out of five. This metric is important to differentiate between two agents with similar final performance but very different LCA.
\begin{table}[t]
  \centering
  \resizebox{0.9\columnwidth}{!}{
  \begin{tabular}{lcccc}
    \toprule
    Environment &TD3 & Static-TD3 & DS-TD3 (ours) & SAC \\
    \midrule
    HalfCheetah-v3 & 1.7686 & 1.7666 & \textbf{1.9560} & 1.7297 \\
    Walker2d-v3 & 0.5264 & 0.5167 & \textbf{0.6956} & 0.6128  \\
    Hopper-v3 & 0.4788 & 0.4984 & 0.5435 & \textbf{0.5572} \\
    Ant-v3 &  0.5524 & 0.5807 & 0.6623 & \textbf{0.7969}\\
    Humanoid-v3 & 0.3635 & 0.5182 & \textbf{0.6089} & 0.5639 \\
    \bottomrule
  \end{tabular}
  }
\caption{Learning curve area (LCA) ($\times$ 5000) of different methods.}
\label{table:learning_speed}
\end{table}
\begin{table*}[t]
  \centering
  \resizebox{1.7\columnwidth}{!}{
  \begin{tabular}{llllll}
    \toprule
    Method & HalfCheetah-v3 &  Walker2d-v3 & Hopper-v3 & Ant-v3 & Humanoid-v3  \\
    \midrule
    TD3 & 11153.48$\pm$473.29 & 4042.36$\pm$576.57  & 2184.78$\pm$1224.14 & 4287.69$\pm$1080.88 &  3809.15$\pm$1053.40\\
    Static-TD3 & 10583.84$\pm$307.03 & 3951.01$\pm$443.78 & 3570.88$\pm$43.71 & 4148.61$\pm$801.34 & 4989.47$\pm$546.32 \\
    DS-TD3 (ours) & \textbf{11459.88$\pm$482.55} & \textbf{4870.57$\pm$525.33}&\textbf{3587.17$\pm$70.62} & 5011.56$\pm$596.95& 5238.16$\pm$121.71  \\
    SAC & 11415.23$\pm$357.22  & 4566.18$\pm$448.25 &  3387.36$\pm$148.73 &\textbf{5848.64$\pm$385.85} & \textbf{5518.61$\pm$97.03} \\
    \bottomrule
  \end{tabular}
  }
\caption{Average return ($R$) over the last 10 evaluations of 1 million time steps.}
\label{final_performance}
\end{table*}
\textbf{Performance.} Table \ref{final_performance} shows the average return ($R$) over the last 10 evaluations. DS-TD3 outperforms TD3 in all environments. Interestingly, it improves TD3 performance by 2.75\%, 20.48\%, 64.18\%, 16.88\%, and 37.51\% on HalfCheetah-v3, Walker2d-v3, Hopper-v3, Ant-v3, and Humanoid-v3 respectively. Static-TD3 has a close performance to TD3 in most cases except for Humanoid-v3, where Static-TD3 outperforms TD3 by 30.98\%. DS-TD3 has a better final performance than SAC in three environments. 

\subsection{Analysis}
\label{sec:analysis}
\textbf{Memory and Computation Costs.} We analyze the costs needed for the training process by calculating the FLOPS and \#params for the actor and critics. We performed this analysis on Half-Cheetah-v3. \#params for dense TD3 is 214784, which requires 1$\times$(1.07e14) FLOPs to train. With our DS-TD3, we can find a much smaller topology that can effectively learn the policy and the function value, achieving higher performance than TD3 with a sparsity level of 51\%. This consequently reduces the number of required FLOPs to 0.49$\times$(1.07e14). 

\begin{figure}[ht]
 \centering
 \begin{subfigure}{0.45\columnwidth}
    \centering
    \includegraphics[width=\linewidth]{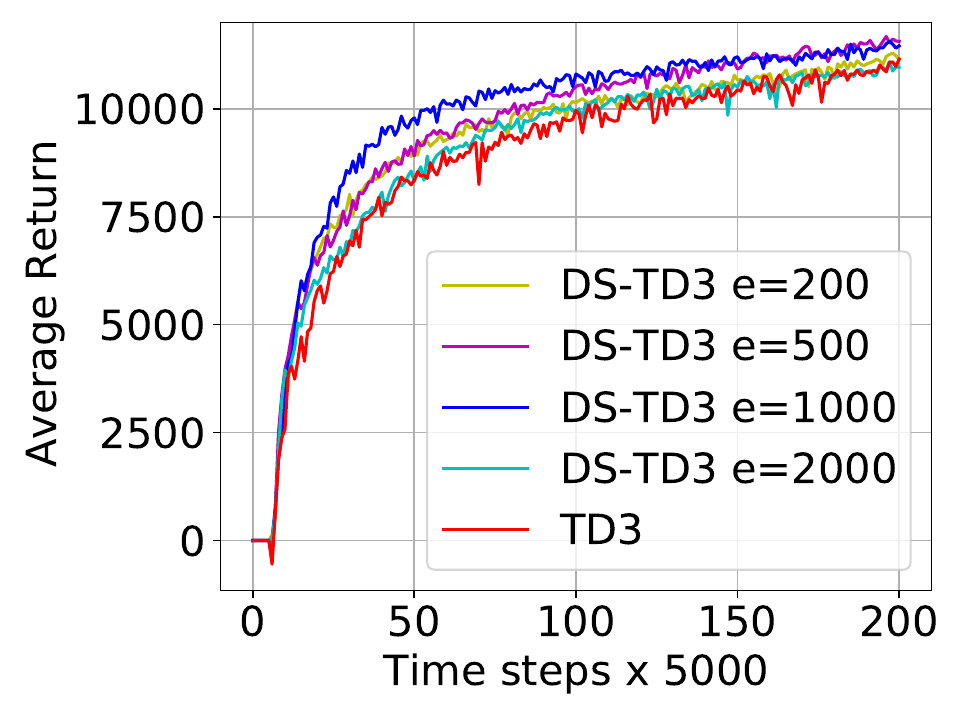}
      \caption{}
      \label{different_schedules}
 \end{subfigure}
 \begin{subfigure}{0.45\columnwidth}
    \centering
    \includegraphics[width=\linewidth]{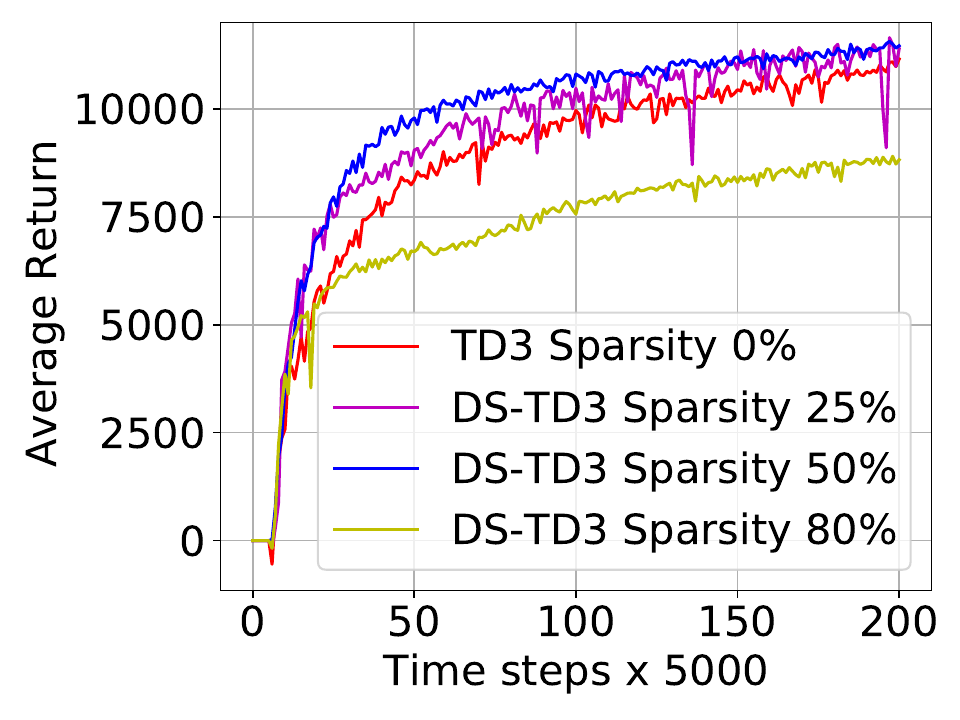}
    \caption{}
    \label{different_sparsity_level}
    \end{subfigure}
\caption{The learning curves of DS-TD3 on HalfCheetah-v3 using different adaptation schedules (a) and sparsity levels (b).}
\end{figure}

\textbf{Adaptation Schedule.} We analyze the effect of the adaptation schedule on the performance. In particular, we ask how frequently the sparse topology should be adapted? We performed this analysis on HalfCheetah-v3. Figure \ref{different_schedules} shows the learning curves of DS-TD3 using different adaptation schedules controlled by the hyperparameter $e$ (Section \ref{method}). Adapting the topology very frequently (i.e., $e\in\{200,500\}$) would not allow the connections to grow and learn in the dynamic changing nature of RL. The current adaptation process could remove some promising newly added connections from the previous adaptation process. This would be caused by a biased estimate of a connection's importance as it becomes a factor of the length of its lifetime. Hence, the very frequent adaptation would increase the chance of replacing some promising topologies. With less frequent adaptation cycles, $e=1000$ (the setting from the paper), DS-TD3 can learn faster and eventually achieves higher performance than other baselines. Giving the connections a chance to learn helps in having better estimates of the importance of the connections. Hence, it enables finding more effective topologies by replacing the least effective connections with ones that better fit the data.
However, increasing the gap between every two consecutive adaptation processes to $2000$ steps decreases the exploration speed of different topologies. As illustrated in the figure, DS-TD3 with $e=2000$ has a close learning behavior and final performance to the dense TD3. Yet, it still offers a substantial reduction in memory and computation costs. This analysis reveals the importance of the adaptation schedule in the success of introducing DST to the DRL field.

\textbf{Sparsity Level.} We analyze the performance of our proposed method using different sparsity levels. Figure \ref{different_sparsity_level} shows the learning curves of the dense TD3 and DS-TD3. By removing 25\% of the connections and training the sparse topology dynamically using DS-TD3, we can achieve a faster learning speed and a performance increase of 2.11\%. More interestingly, with a higher reduction in the size of the networks by 50\%, we achieve a much faster learning speed. However, when the network has a very high sparsity level (i.e., 80\%), it fails to learn effective representations for the reinforcement learning setting. Learning DRL agents using very high sparse networks is still an open-challenging task.

\textbf{Learning Behavior and Speed.} DRL agents learn through trial-and-error due to the lack of true labels. An agent starts training with samples generated from a purely exploratory policy, and new samples are drawn from the learning policy over time. Our results show that dynamic sparse agents have faster adaptability to the newly improved samples, thanks to the generalization ability of sparse neural networks~\cite{hoefler2021sparsity}. This leads to higher learning speed, especially at the beginning of the training. We hypothesize that dense neural networks, being over-parameterized, are more prone to  memorize and overfit the inaccurate samples. A longer time is required to adapt to the newly added samples by the improved policy and forget the old ones.

\begin{figure}
 \centering
 \begin{subfigure}{0.45\columnwidth}
    \centering
    \includegraphics[width=\linewidth]{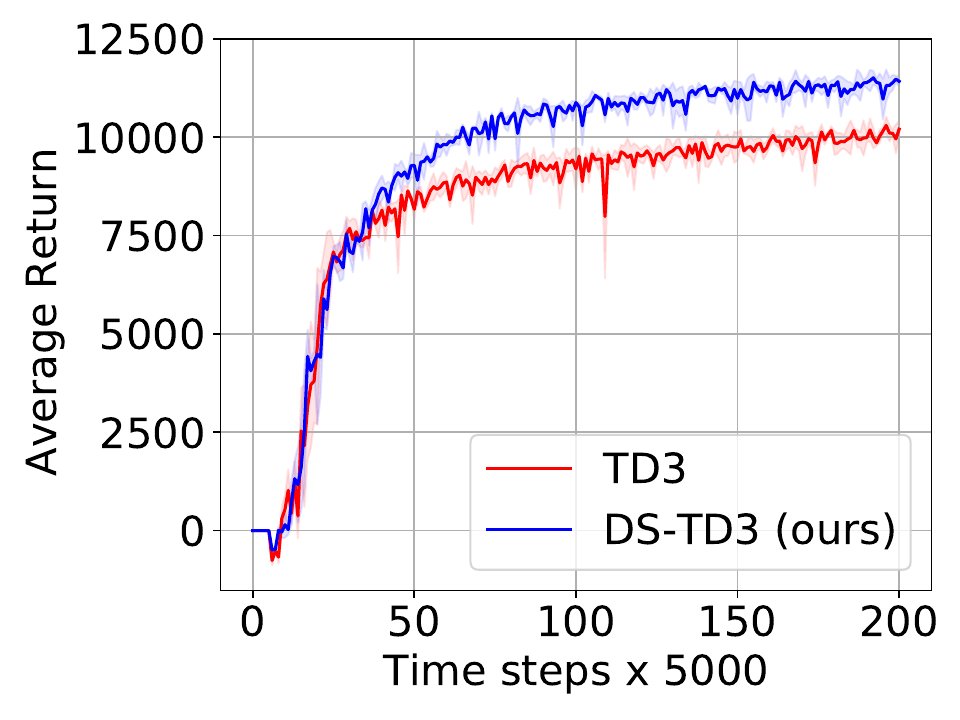}
      \caption{}
      \label{initial_50}
 \end{subfigure}
 \begin{subfigure}{0.45\columnwidth}
    \centering
    \includegraphics[width=\linewidth]{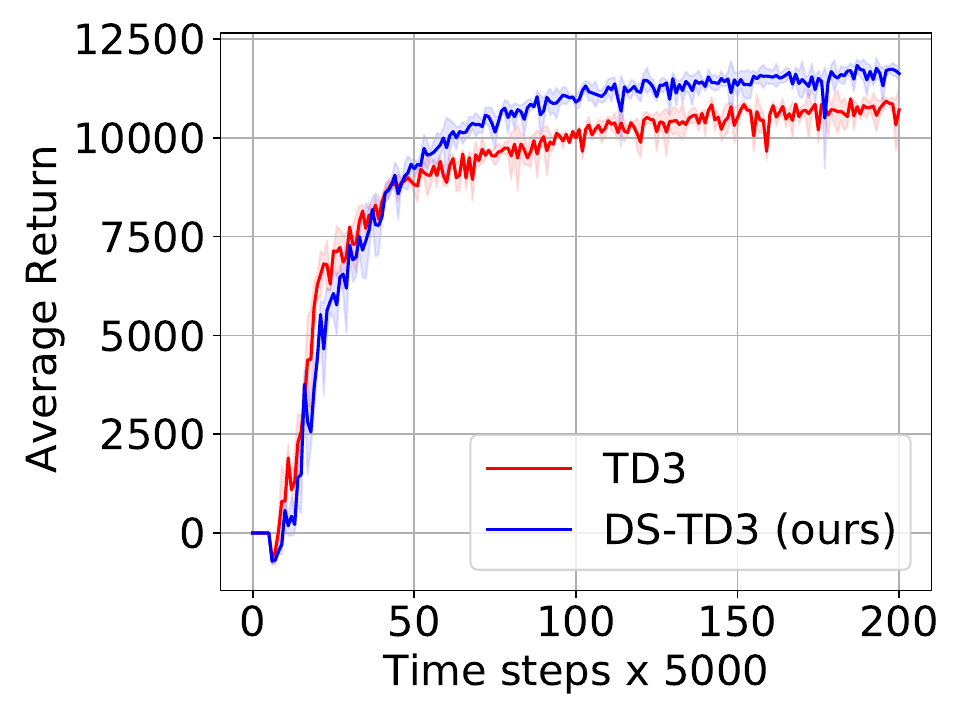}
      \caption{}
      \label{initial_70}
 \end{subfigure}
 \caption{Learning curves of agents that start training with samples drawn from policies trained for $5\times10^5$ (a) and $7\times10^5$ steps (b). }
 \label{initialbuffer}
\end{figure}

To validate this hypothesis, we analyze the behavior of a dense TD3 agent when it starts training with samples generated from a learned policy instead of a purely exploratory one. We use two learned policies trained for $5\times10^5$ and $7\times10^5$ steps to draw the initial samples (see Appendix \ref{appendix:learningspeed}). We performed this experiment on HalfCheetah-v3. As illustrated in Figure \ref{initialbuffer}, the learning speed of DS-TD3 and TD3 becomes close to each other at the beginning. Afterward, DS-TD3 performs better than TD3 since the new samples are generated from the current learning policies. With initial samples drawn from more improved policy (Figure \ref{initial_70}), dense TD3 learns faster. It achieves higher performance than the baseline that starts learning with samples drawn from the policy trained for $5\times10^5$ steps (Figure \ref{initial_50}). On the other hand, DS-TD3 is more robust to over-fitting, less affected by the initial samples, and quickly adapt to the improved ones over time.     

% performed additional experiments in which we start training randomly initialized networks with an initial buffer filled with samples generated from a learned policy instead of a purely exploratory one.  In particular, we trained TD3 and DS-TD3 for 500000 steps on HalfCheetah-v3. Then, we use the learned policies to initialize the buffers for learning new agents based on TD3 and DS-TD3 from scratch on the same environment. We compare these models to the models that start training from an initial buffer filled with samples generated from a random policy (the typical practice). 

% We observe that when the initial buffers are filled with samples generated from the learned policies, the learning speed of DS-TD3 and TD3 are close to each other at the beginning. Afterward, DS-TD3 starts to perform better than TD3 since the new samples are generated from the current learning policies. 
% We believe that the generalization ability [2], robustness to over-fitting, and the dynamic optimization of the sparse topology to fit and adapt to the incoming data contribute to the higher performance achieved by the sparse models.

\section{Conclusion}
Introducing dynamic sparse training principles to the deep reinforcement learning field provides an efficient training process for DRL agents. Our dynamic sparse agents achieve higher performance than the state-of-the-art methods while reducing the memory and computation costs by 50\%. Optimizing the sparse topology during training to adapt to the incoming data increases the learning speed. Our findings show the potential of dynamic sparse training in advancing the DRL field. This would open the path to efficient DRL agents that could be trained and deployed on low-resource devices where memory and computation are strictly constrained.

\section{Acknowledgments}
This material is partly based upon work supported by the Google
Cloud Research Credits program.
%% The file named.bst is a bibliography style file for BibTeX 0.99c
\bibliographystyle{named}
\bibliography{ijcai22}
\newpage
\clearpage
\appendix
\section{Experimental Details}
\label{appendix:Experimental_Details}
For a direct comparison with the TD3 algorithm, we follow the same setting as in \cite{fujimoto2018addressing}. We use multi-layer perceptrons for the actor and critics networks with two hidden layers of 256 neurons and a ReLU activation function. A Tanh activation is applied to the output layer of the actor network. Sparse connections are allocated in the first two layers for all networks while the output layer is dense. We use $\lambda^{2}$ of 64 for all environments. In contrast, $\lambda^{1}$ varies across environments because it depends on each environment's state and action dimensions. We use $\lambda^{1}$ of 7 for HalfCheetah-v3, Hopper-v3, and Walker2d-v3. For Ant-v3 and Humanoid-v3, we use $\lambda^{1}$ of 40 and 61, respectively. The same sparsity levels are used for Static-TD3. %These values are used for both actor and critic networks. %$\epsilon_{1}$ varies across environments because it depends on the state and action dimension of each environment. Table \ref{epsilon1_values} shows the values selected for each of the environments we evaluated on. These values are used for both actor and critic networks. 

We adapt the sparse connections every $e$ time steps, with $e=1000$. A fraction of the sparse connections $\eta$ is adapted with $\eta=0.05$. The networks are trained using Adam optimizer with a learning rate of 0.001 and a weight decay of 0.0002. The networks are trained with mini-batches ($N$) of 100, sampled uniformly from a replay buffer containing the entire history of the agent.

Following the TD3 algorithm \cite{fujimoto2018addressing}, we added noise of $\epsilon$ $\sim$ $\mathcal{N}(0 ,0.2)$  to the actions chosen by the target actor network and clipped to $(-0.5, 0.5)$. The actor and target networks are updated every 2 steps $(d = 2)$. The $\tau$ used for updating the target networks equals 0.005.  A purely exploratory policy is used for the first 25000 time steps, then an off-policy exploration strategy is used with Gaussian noise of $\mathcal{N}(0, 0.1)$ added to each action.

The hyperparameters for the dynamic sparse training ($\lambda^{1}, \lambda^{2}, \eta, e$) are selected using random search. Each environment is run for 1 million time steps with evaluations every 5000 time steps, where each evaluation reports the average return over 10 episodes with no exploration noise. LCA is calculated using the average return computed every 5000 time steps. Our results are reported over 5 seeds. 

All models are implemented with PyTorch and trained on Nvidia GPUs. We use the official code from the authors of TD3 \cite{fujimoto2018addressing}, which has an MIT license, to reproduce the results of TD3 with the above settings. 

For SAC, we use the Pytorch implementation from \cite{pranz24}. We follow the settings from the original paper \cite{haarnoja2018soft} with the same architecture used for TD3. The networks are trained using Adam optimizer with a learning rate of 0.0003 and mini-batches of 256. We use $\tau$ of $0.005$ and a target update interval of 1. The models are trained for 1M steps.

\section {Evaluation Metrics}
\label{appendix:evaluation_metrics}
In this appendix, we explain the details of the metrics used to assess the performance of our proposed method. 

\textbf{Return ($\textbf{R}$).} The average return is the standard metric used in the RL research to measure the \textit{performance} of an agent. The return is the sum of rewards ($r$) obtained in one episode of $T$ steps. $R$ is calculated as follows:
\begin{equation}
    R = \sum_{t=1}^{T} r_t.
\end{equation}

\textbf{Learning curve area (LCA).} This metric estimates the learning speed of a model. LCA measures the area under the training curve of a method. Intuitively, the higher learning curve, the faster the learner is. We adapt this metric from \cite{chaudhry2018efficient} to fit the reinforcement learning paradigm. LCA is calculated as follows:   
\begin{equation}
    LCA = \frac{1}{\Delta}\int_{0}^{\Delta} R(t) dt = \frac{1}{\Delta}  \sum_{t=0}^\Delta R(t),
\end{equation}
where $\Delta$ is the number of training steps and $R$ is the average return.  

\textbf{Network size (\#params).} This metric estimates the \textit{memory cost} consumed by an agent. The network size is estimated by the summation of the number of connections allocated in its layers as follows:
\begin{equation}
    \#params = \sum_{l=1}^{L} \norm{\bm{W}^{l}}_{0},
\end{equation}
where $\bm{W}^{l}$ is the actual weights used in layer $l$, $\norm{.}_0$ is the standard $L_0$ norm, and $L$ is the number of layers in the model. For sparse neural networks, $\norm{\bm{W}^{l}}_{0}$ is controlled by its defined sparsity level for the model.

\textbf{Floating-point operations (FLOPs).} This metric estimates the \textit{computational cost} of a method by calculating how many FLOPs are required for training. We follow the method described in \cite{evci2020rigging} to calculate the FLOPs. The FLOPs are calculated with the total number of multiplications and additions layer by layer in the network. 

FLOPs is the typical used metric in the literature to compare a DST method against its dense counterpart. The motivation is twofold. First, it gives an unbiased estimate of the actual required number of operations since the running time would differ from one implementation to another. Second, more importantly, existing dynamic sparse training methods in the literature are currently prototyped using masks over dense weights to simulate sparsity \cite{hoefler2021sparsity}. This is because most deep learning specialized hardware is optimized for dense matrix operations. Therefore, the running time using these prototypes would not reflect the actual gain in memory and speed using a truly sparse network. Therefore, the FLOPs and network parameters are the current commonly used metrics to estimate the computation and memory costs respectively for sparse neural networks \cite{hoefler2021sparsity}. 

\section{Hardware and Software Support}
\label{appendix:HW_SW}
 As a joint community effort, research on sparsity is going into three parallel directions: First, hardware that supports sparsity. NVIDIA released NVIDIA A100, which supports a 50\% fixed sparsity level \cite{zhou2020learning}. Second, software libraries that support truly sparse implementations. Efforts have been started to be devoted to supervised learning \cite{onemillionneurons}. Third, algorithmic methods, our focus, that aim to provide approaches that achieve the same performance of dense models using sparse networks \cite{hoefler2021sparsity}. With the parallel efforts in the three directions, we would be able to actually provide faster, memory-efficient, and energy-efficient deep neural networks. This is further discussed in \cite{hooker2021hardware,mocanu2021sparse}.

\section{Learning Behavior Analysis}
\label{appendix:learningspeed}
To analyze the effect of the absence of true labels in the behavior of dense and dynamic sparse agents, we perform experiments in which an agent starts learning from samples drawn from a learned policy (Section \ref{sec:analysis}). We test two learned policies with different performance to study how the quality of the initial samples affects the learning behavior. To this end, we train two dense policies using TD3 for $5\times10^5$ and $7\times10^5$ steps on Half-Cheetah-v3. Similarly, we train two sparse policies for the same time steps using DS-TD3. Instead of using a purely exploratory policy, we draw samples from the learned policies to fill the initial buffers for dense and dynamic sparse agents that learn from scratch. 

As discussed in Section \ref{sec:analysis}, the performance of dense DRL agents is more affected by the initial samples. The better the samples are, the higher performance is. On the other hand, dynamic sparse agents adapt quickly to the improving samples over time and are less affected by the quality of the initial samples.

\section{DS-SAC}
\label{appendix:DS-SAC}
In this appendix, we demonstrate that our proposed dynamic sparse training approach can be integrated with other state-of-the-art DRL methods. We use the soft actor-critic (SAC) method \cite{haarnoja2018softapp} and name our improved version of it as Dynamic Sparse SAC or DS-SAC. 

SAC is an off-policy algorithm that optimizes a stochastic policy. A key feature of this method is entropy regularization. The policy is trained to maximize a trade-off between expected return and entropy (a measure of randomness). Thus, the agent addresses the exploration-exploitation trade-off, which results in policies that explore better. Algorithm \ref{alg:DS_soft_actor_critic} shows our proposed DS-SAC. We integrated the four components of our approach (sparse topology initialization, adaptation schedule, topological adaptation, and maintain sparsity levels) into the original algorithm. 

\textbf{Tasks.} We compared our proposed DS-SAC with SAC. We performed our experiments on five MuJoCo control tasks. Namely, we tested the following environments: HalfCheetah-v3, Hopper-v3, Walker2d-v3, Ant-v3, and Humanoid-v3.

\textbf{Experimental settings.} We follow the setting from the SAC method \cite{haarnoja2018soft}. All networks are multilayer perceptrons with two hidden layers of 256 neurons and a ReLU activation function. The networks are training with Adam optimizer and a learning rate of 0.0003. We use mini-batches of 256. Each environment is run for 1 million steps. As DRL algorithms and their variants behave differently in various settings/environments, to cover a wider range of possible scenarios, we study here the case of hard target update where $\tau=1$ \cite{haarnoja2018soft}. Following the original paper, target update interval= 1000. We use a  temperature $\alpha$ of 0.2. Table \ref{table:sparsity_level} shows the value used for $\lambda^1$ and $\lambda^2$ to determine the sparsity levels for DS-SAC. We use $e$ of 1000 and $\eta$ of 0.1. The hyperparameters are selected using a random search. Our results are reported over 5 seeds.

\textbf{Metrics.} We used the same metrics discussed in Section \ref{sec:experiments} to assess the performance of our proposed method.
\begin{algorithm}[H]
\caption{DS-SAC}
\label{alg:DS_soft_actor_critic}
\begin{algorithmic}[1]
\STATE Require: $\lambda^{l}$, $\eta$, $e$
\STATE Create $\bm{M_{\phi}}$, $\bm{M_{\theta_1}}$, and $\bm{M_{\theta_2}}$ with Erdős–Rényi random graph with sparsity level $\lambda^l$
\STATE $\theta_{1} \leftarrow$ $\theta_{1} \odot$ $\bm{M_{\theta_1}}$, $\theta_{2} \leftarrow$ $\theta_{2} \odot$ $\bm{M_{\theta_2}}$, $\phi \leftarrow \phi \odot \bm{M_{\phi}}$
\STATE Initialize target networks $\bar\theta_1 \leftarrow \theta_1$, $\bar\theta_2 \leftarrow \theta_2$
\STATE $\mathcal{D}\leftarrow\emptyset$ // Initialize an empty replay pool
\FOR{each iteration}
	\FOR{each environment step}
	    \STATE $a_t \sim \pi_\phi(a_t|s_t)$ // Sample action from the policy
	    \STATE $s_{t+1} \sim p(s_{t+1}| s_t, a_t)$ // Sample transition from the environment
	    \STATE $\mathcal{D} \leftarrow \mathcal{D} \cup \left\{(s_t, a_t, r(s_t, a_t), s_{t+1})\right\}$ // Store the transition in the replay pool
	\ENDFOR
	\FOR{each gradient step}
	    \STATE $\theta_i \leftarrow \theta_i - \lambda_Q \hat \nabla_{\theta_i} J_Q(\theta_i)$ // Update Q-functions
        \IF{$t$ mod $e$} 
          \STATE $\theta_i\leftarrow$ TopologicalAdaptation$(\theta_i, \bm{M_{\theta_{i}}},\eta)$ (Algo. \ref{alg:topo_evolution})
       \ENDIF
        
	    \STATE $\phi \leftarrow \phi - \lambda_\pi \hat \nabla_\phi J_\pi(\phi)$ // Update policy weights
       \IF{$t$ mod $e$}
        \STATE $\phi \leftarrow$ TopologicalAdaptation$(\phi, \bm{M}_{\phi},\eta)$ (Algo. \ref{alg:topo_evolution})
       \ENDIF
	   % \STATE $\alpha \leftarrow \alpha - \lambda \hat \nabla_\alpha J(\alpha)$ $\rhd$ Adjust temperature
	    \STATE $\bar\theta_i\leftarrow \tau \theta_i + (1-\tau)\bar\theta_i$  // Update target network 
	   \STATE $\bar\theta_i \leftarrow$  MaintainSparsity$(\bar\theta_i,\norm{\theta_i}_0)$ (Algo.  \ref{alg:maint_sparsity})
	\ENDFOR
\ENDFOR
\end{algorithmic}
\end{algorithm}

\begin{figure*}
 \centering
 \begin{subfigure}{0.55\columnwidth}
    \centering
    \includegraphics[width=\linewidth]{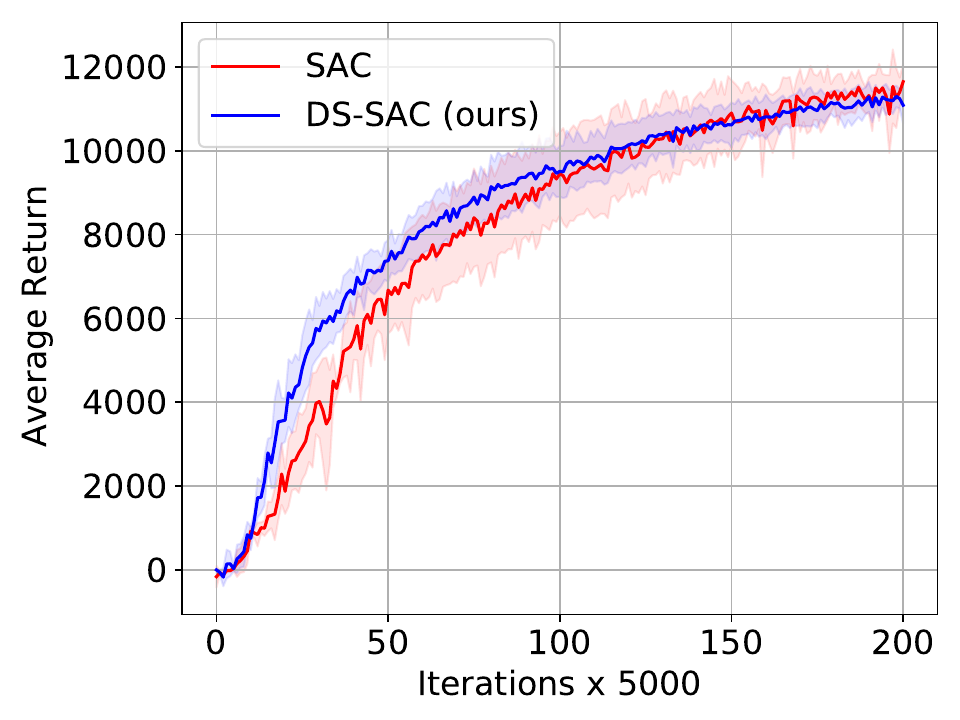}
      \caption{HalfCheetah-v3.}
 \end{subfigure}
 \begin{subfigure}{0.55\columnwidth}
    \centering
    \includegraphics[width=\linewidth]{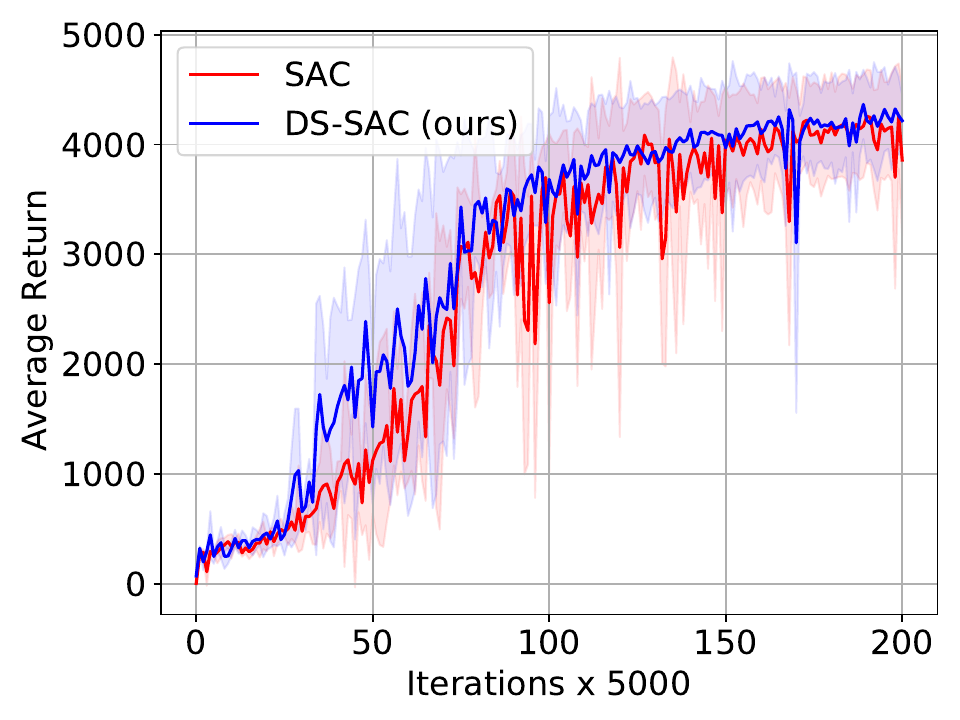}
      \caption{Walker2d-v3.}
 \end{subfigure}
\begin{subfigure}{0.55\columnwidth}
    \centering
    \includegraphics[width=\linewidth]{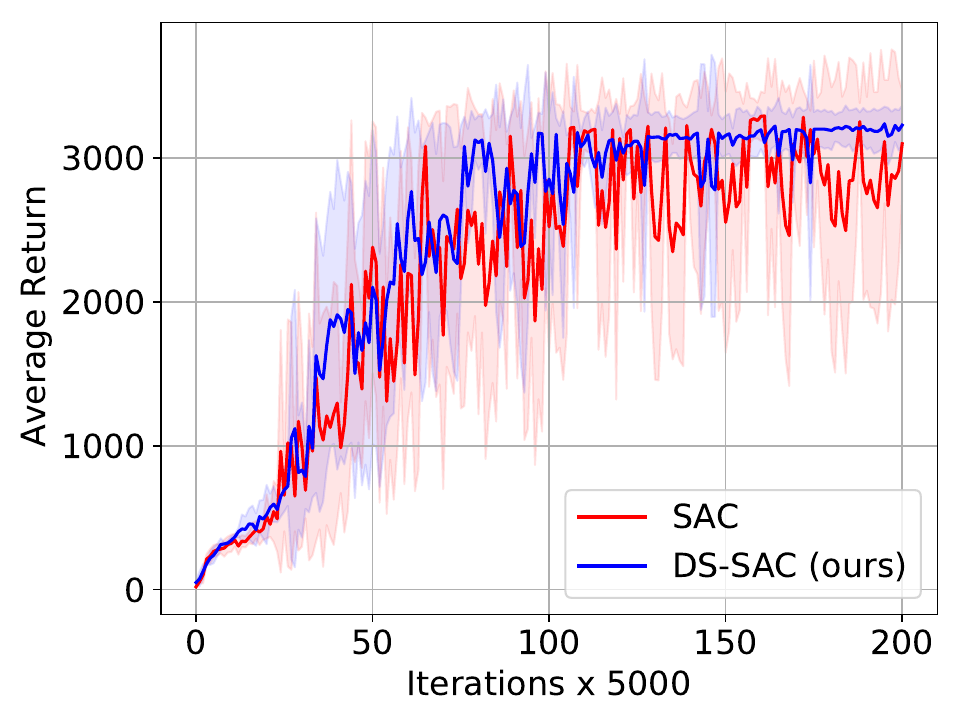}
      \caption{Hopper-v3.}
 \end{subfigure}
 \begin{subfigure}{0.55\columnwidth}
    \centering
    \includegraphics[width=\linewidth]{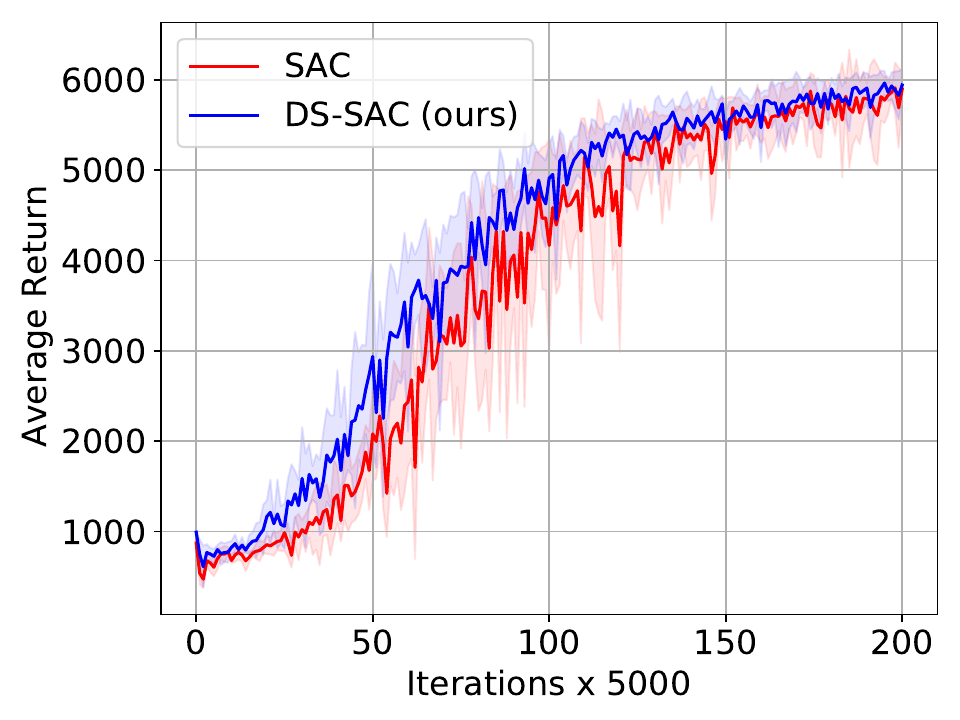}
      \caption{Ant-v3.}
 \end{subfigure}
  \begin{subfigure}{0.55\columnwidth}
    \centering
    \includegraphics[width=\linewidth]{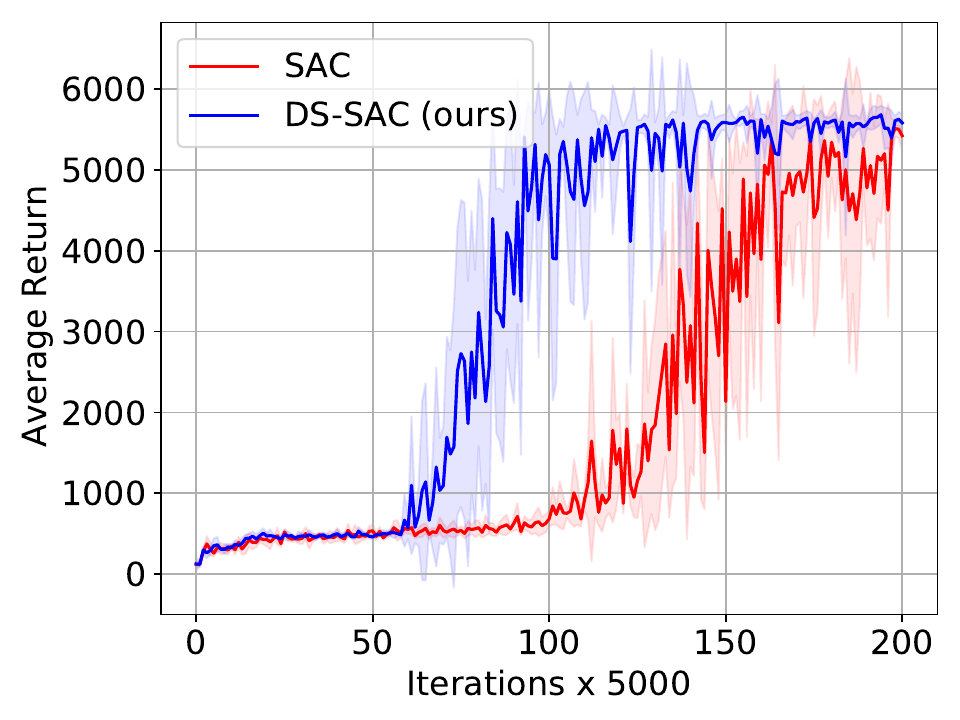}
      \caption{Humanoid-v3.}
 \end{subfigure}
\caption{Learning curves of SAC and DS-SAC on different continuous control tasks. The shaded region represents the standard deviation of the average evaluation over 5 runs.} 
\label{learning_curves_sac}
\end{figure*}

\begin{table}[H]
  \caption{The value used for $\lambda^1$ and $\lambda^2$ in each environment for the DS-SAC algorithm.}
  \label{table:sparsity_level}
  \centering
  %\resizebox{\columnwidth}{!}{
  \begin{tabular}{lcc}
    \toprule
    Environment & $\lambda^1$ & $\lambda^2$  \\
    \midrule
    HalfCheetah-v3 & 12 & 80 \\
    Walker2d-v3 &  12 & 80\\
    Hopper-v3 &  7 &20 \\
    Ant-v3 & 30 & 64\\
    Humanoid-v3 & 61 & 64\\
    \bottomrule
  \end{tabular}
  %}
\end{table}
\textbf{Results.} Figure \ref{learning_curves_sac} shows the learning behavior of DS-SAC and SAC. Consistent with our previous observations, DS-SAC learns faster, especially at the beginning of the training. The LCA of DS-SAC is higher than SAC for all environments, as shown in Table \ref{table:learning_speed_sac}. DS-SAC outperforms the final performance of SAC for all environments except one where it achieves a very close performance to it, as illustrated in Table \ref{final_performance_sac}. Please note that the results of SAC are slightly different from the ones obtained in Section \ref{sec:results} as we study here the hard target update case of SAC \cite{haarnoja2018softapp}. 

These experiments reveal that we can achieve gain in a DRL agent's learning speed and performance while reducing its required memory and computation costs for training. 

\begin{table}[b]
  \caption{Learning curve area (LCA) ($\times$ 5000) of SAC and DS-SAC.}
  \label{table:learning_speed_sac}
  \centering
  %\resizebox{\columnwidth}{!}{
  \begin{tabular}{lcc}
    \toprule
    Environment &SAC & DS-SAC (ours)  \\
    \midrule
    HalfCheetah-v3 & 1.6229 & \textbf{1.7081}\\
    Walker2d-v3 & 0.5368 & \textbf{0.5906} \\
    Hopper-v3 & 0.4441 & \textbf{0.4875} \\
    Ant-v3 & 0.7504 & \textbf{0.8229}\\
    Humanoid-v3 & 0.3776 & \textbf{0.6777}\\
    \bottomrule
  \end{tabular}
  %}
\end{table}

\begin{table}[b]
  \caption{Average return over the last 10 evaluations of 1 million time steps using SAC and DS-SAC.}
  \label{final_performance_sac}
  \centering
  %\resizebox{\columnwidth}{!}{
  \begin{tabular}{lcc}
    \toprule
    Environment &SAC &  DS-SAC (ours) \\
    \midrule
    HalfCheetah-v3 & \textbf{11645.12 $\pm$ 425.585} &11084.39 $\pm$ 445.15\\
    Walker2d-v3 & 3858.20 $\pm$ 689.913 & \textbf{4216.77 $\pm$ 236.23} \\
    Hopper-v3 & 3100.39 $\pm$ 374.45 & \textbf{3229.39 $\pm$ 135.82} \\
    Ant-v3 & 5899.30 $\pm$ 197.15 & \textbf{5943.54 $\pm$ 169.95}\\
    Humanoid-v3 & 5425.56 $\pm$ 196.33 & \textbf{5584.64 $\pm$ 109.40}\\
    \bottomrule
  \end{tabular}
  %}
\end{table}

\end{document}